# Human-Centered LLM-Agent System for Detecting Anomalous Digital Asset Transactions

**Written by Gyuyeon Na**[1]
**Minjung Park**[2]
**Hyeonjeong Cha**[1]
**Sangmi Chai**[1,3,*]
[1]AI and Business Analytics, Ewha Womans University, Seoul, Republic of Korea
[2]Department of Business Administration, Kumoh National Institute of Technology, Gumi, Republic of Korea
[3]Coretrustlink, Seoul, Republic of Korea
amy-na@ewha.ac.kr, mjpark@kumoh.ac.kr , hyeonjeong.cha@ewha.ac.kr

### Abstract

We present HCLA, a human-centered multi-agent system for anomaly detection in digital-asset transactions. The system links three roles—Parsing, Detection, and Explanation—into a conversational workflow that lets non-experts ask questions in natural language, inspect structured analytics, and obtain context-aware rationales. Implemented with an open-source web UI, HCLA translates user intents into a schema for a classical detector (XGBoost in our prototype) and returns narrative explanations grounded in the underlying features. On a labeled Bitcoin-mixing dataset (Wasabi Wallet, 2020–2024), the baseline detector reaches strong accuracy while HCLA adds interpretability and interactive refinement. We describe the architecture, interaction loop, dataset, evaluation protocol, and limitations, and we discuss how a human-in-the-loop design improves transparency and trust in financial forensics

## Introduction

Decentralized-finance ecosystems complicate risk monitoring: mixers and privacy tools obscure provenance, and detectors that perform well often behave as opaque systems that require expertise to operate. What users need is a surface where they can ask, probe, and verify.

HCLA addresses this need by orchestrating three agents in one dialogue: a Parsing LLM converts everyday re- quests into machine-readable variables; a Detection model scores transactions; an Explanation LLM converts scores and features into reasons a person can audit. The core idea is simple: let people steer anomaly analysis without mastering pipelines, and make each step auditable end-to- end. (See Figure 1 and Figure 2 for the interaction; Table tab:relatedwork contrasts related systems.) Contributions.

A modular, conversational workflow that couples natural-language intents with structured anomaly analytics; A clear separation of concerns—Parsing / Detection / Explanation—so improvements in one module do not destabilize others; Evidence that interactive, human-readable rationales improve accessibility and trust without sacrificing detector performance.

*Corresponding author: smchai@ewha.ac.kr

## Related Work

Research has explored

(i) rules and supervised detectors that require labeled data and expert operation;

(ii) LLMs for parsing or summarizing events; and

(iii) explanation methods that verbalize model outputs. Prior LLM systems typically stop at one-way reporting or run as a single agent without a mechanism for users to refine reasoning. Table tab:relatedwork summarizes representative approaches and their limitations that HCLA targets: sustained dialogue, modularity, and reproducibility under prompt contracts.

Recent studies have applied large language models (LLMs) to anomaly detection in various domains, such as finance, cybersecurity, and blockchain analytics. Recent studies have applied large language models (LLMs) to anomaly detection in various domains, such as finance, cybersecurity, and blockchain analytics (Tsai et al. 2025; Park 2024; He et al. 2025; Lin et al. 2025; Yu et al. 2025; Li et al. 2025; Watson 2025; Lei et al. 2025; Sun et al. 2025; Jia et al. 2025).

## System Overview: HCLA Framework

HCLA operationalizes anomaly detection as a conversational workflow between human users and AI agents. Implemented through Gradio, the interface allows natural-language queries such as “Analyze transactions from my wallet over the past week.”

- The Parsing Agent (ChatGPT) converts free-form inputs into structured JSON schemas.
- The Detection Agent (XGBoost) computes anomaly probabilities using temporal, transactional, and graph-connectivity features.
- The Explanation Agent (Gemini) translates numeric scores into plain-language reasoning, supporting iterative questioning (“Why is this transaction high-risk?”).

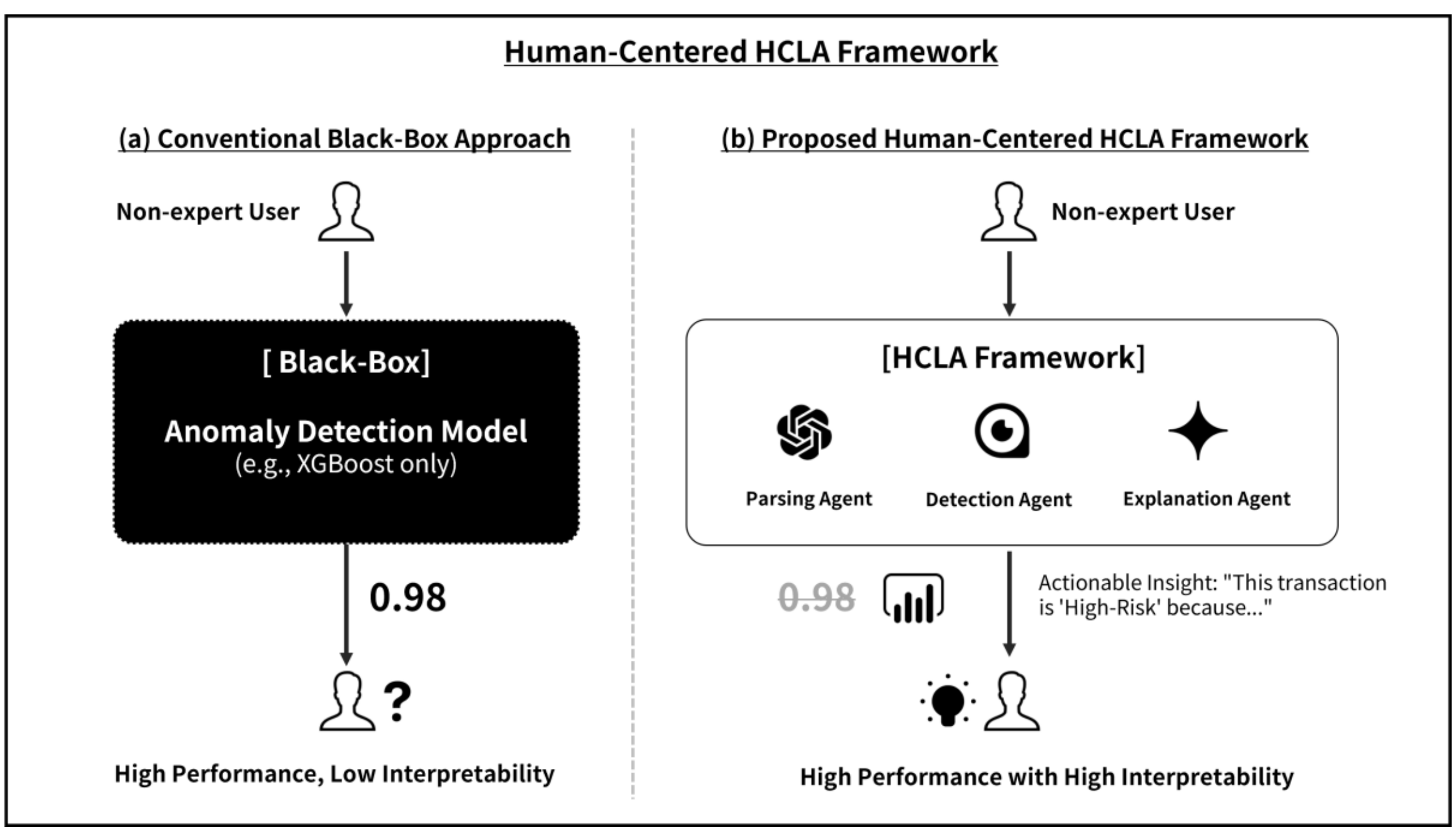

Figure 1: Bridging the Interpretability Gap: From a Black-Box Model to the Human-Centered HCLA Framework.

Agents communicate via JSON messages under a central LLM controller that preserves semantic alignment between parsed features, model outputs, and generated explanations. This architecture ensures transparency, traceability, and adaptability—core attributes of human-centered AI.

HCLA treats anomaly detection as conversation plus computation.Architecture Parsing Agent (LLM-A). Extracts entities and constraints (time windows, wallet/cluster IDs, value ranges) and emits a normalized JSON schema that the detector consumes. Ambiguities are clarified interactively. (Figure 1.)

Detection Agent (XGBoost baseline). Consumes temporal and transactional features (frequency, direction, connectivity) and outputs anomaly probabilities. The interface allows drop-in alternatives (e.g., GNNs, temporal models) without changing orchestration.

Explanation Agent (LLM-B). Maps scores and features to concise, context-aware narratives and supports follow-up questions ("Why is this high-risk?"). (Figure 2 .) Agents exchange JSON under a lightweight controller that preserves semantic alignment across steps.Interaction Loop

- User query → natural language (e.g., "Past week, my wallet—anything suspicious?").
- Parsing → JSON schema (wallet, range, entities). (See Listing 1–Listing 2)
- Detection → per-transaction anomaly likelihoods.
- Explanation → reasons tied to features ("repeated high-value transfers to unverified counterparties during off-peak hours").
- Refinement → users filter, group, or pivot (e.g., "only exchange-linked clusters"), and the loop continues.

Listing 1: User query consumed by the Parsing Agent.

```
1  On September 20, 2025, at 11:00 PM (UTC
      +9), I received 0.8 BTC (worth about
      $51,200) to my address 1A2b3C from
      the counterparty bc1qxxx.    Please
      check if this transaction looks
      suspicious.
```

Listing 2: Parsing Agent output as structured JSON schema.

```
{
  "Date": "2025-09-20T23:00:00+09:00",
  "Receiving Address": "1A2b3C...",
  "Counterparty Address": "bc1qxxx",
  "Value": 0.8,
  "USD Value": 51200.0
}
```

## Methodology

This section describes the methodological foundation of the proposed HCLA system. It covers the dataset characteristics, model integration strategy between LLM agents and the underlying anomaly-detection algorithm, and the evaluation metrics used to assess both performance and interpretability.

### Dataset

We use a labeled Bitcoin-mixing dataset derived from Wasabi Wallet (2020–2024): 318,388 normal and 69,031 anomalous transactions with hashed IDs, sender/receiver, USD values, and temporal fields. We train on 2020–2022 and test on 2023–2024 to emulate deployment. Records missing key fields are dropped; extreme values are kept to preserve anomalies.

Table 1: Comparison of LLM-based Anomaly-Detection Approaches

| Study | Summary | Limitation |
|---|---|---|
| *RAAD-LLM: Adaptive Anomaly Detection Using LLMs and RAG* (Russell-Gilbert 2025) | LLM retrieves relevant context via RAG when users issue natural-language anomaly queries and summarizes results. | Handles input only at the "event description" level; lacks sustained, conversational interaction and feedback incorporation. |
| *LLM-Augmented Explanations for Graph-Based Crypto Anomaly Detection* (Watson 2024) | LLM reformulates outputs from graph-based models into human-readable explanations for end users. | One-way reporting; users cannot interactively explore or refine anomaly reasoning. |
| *CALM: Continuous, Adaptive, and LLM-Mediated Anomaly Detection in Time-Series Streams* (Devireddy and Huang 2025) | LLM acts as a judge for time-series anomaly detectors, offering adaptive summaries and continuous updates. | System-centric design focused on model performance, not user interaction or interpretability. |
| *AnoLLM: Large Language Models for Tabular Anomaly Detection* (Tsai et al. 2025) | Tabular records are serialized into prompts and analyzed by LLMs for descriptive anomaly outputs. | Provides explanations but lacks conversational interface and modular structure (Parsing–Detection–Explanation). |
| *Anomaly Detection for Short Texts* | Users paste logs or short texts; LLM summarizes cause and detection results conversationally. | Single-agent pipeline; no explicit modular separation, limiting extensibility and control. |
| *HCLA (proposed framework)* | HCLA integrating Parsing, Detection, and Explainer agents through Gradio, stabilized by prompt engineering. | Enables conversational interaction, modular reproducibility, and transparent reasoning for non-expert users. |

## Model Integration

The analytical backbone uses XGBoost, integrated with LLM-driven orchestration(**Figure 3**):

Parsing Feature Mapping – Natural-language queries are mapped to structured schema via prompt-templated extraction (wallet, time range, amount). Detection Pipeline – Engineered temporal-transactional features feed the XGBoost model, producing anomaly probabilities p(a). LLM-Assisted Interpretation – The Explanation Agent converts scores into narrative justifications, e.g., "This transaction shows a 0.84 anomaly score due to repeated high-value transfers to unverified counterparties during off-peak hours."

Interactive Loop – Users pose follow-up queries; the Parsing Agent updates context, re-triggering detection and explanation.

**1. Parsing and Feature Mapping.** User input is processed by the Parsing Agent (ChatGPT) via a structured prompt template:

> "Extract wallet address, time range, and transaction details from the following description and output a JSON schema."

This ensures deterministic schema construction, resolves missing values through contextual reasoning, and maps linguistic expressions ("past week", "high-value transfers") to numerical features (e.g., day range=7, value > threshold).

**2. Detection Pipeline.** Parsed features are passed to an XGBoost model trained on engineered temporal-transactional attributes. The model outputs an anomaly probability $p(a_i)$ for each transaction $i$. Although XGBoost serves as the baseline in this prototype, the interface supports substitution with graph-based or temporal models (e.g., GCN-GRU) without modifying the agent layer. This modularity decouples learning algorithms from interaction logic, enabling rapid experimentation.

**3. LLM-Assisted Interpretation.** Once the model produces a score vector, the Explanation Agent (Gemini) translates the numerical outputs into natural-language interpretations. Prompt engineering constrains the response style—for example, "concise risk summary with probabilistic rationale." A typical explanation might read:

> "This transaction shows a high anomaly score (0.84) due to repeated transfers exceeding the 95th-percentile value to unverified counterparties during off-peak hours."

Users can then follow up with additional queries, and the Parsing Agent re-initiates the workflow with preserved conversational context. This cyclical structure embodies the **human-in-the-loop** principle that distinguishes HCLA from conventional, one-shot detectors.

**4. API and System Integration.** Agents communicate through lightweight JSON messages within a Gradio interface. The central controller manages context buffers and ensures that inputs, outputs, and intermediate explanations remain synchronized. All modules are stateless between user sessions but maintain intra-session memory for multi-turn coherence. This design achieves real-time responsiveness while preserving interpretability and traceability.

## Metrics Protocol

We report Accuracy, Precision, Recall, F1, and latency for the detector plus qualitative judgments of explanation clarity. Using fixed random splits and version-controlled

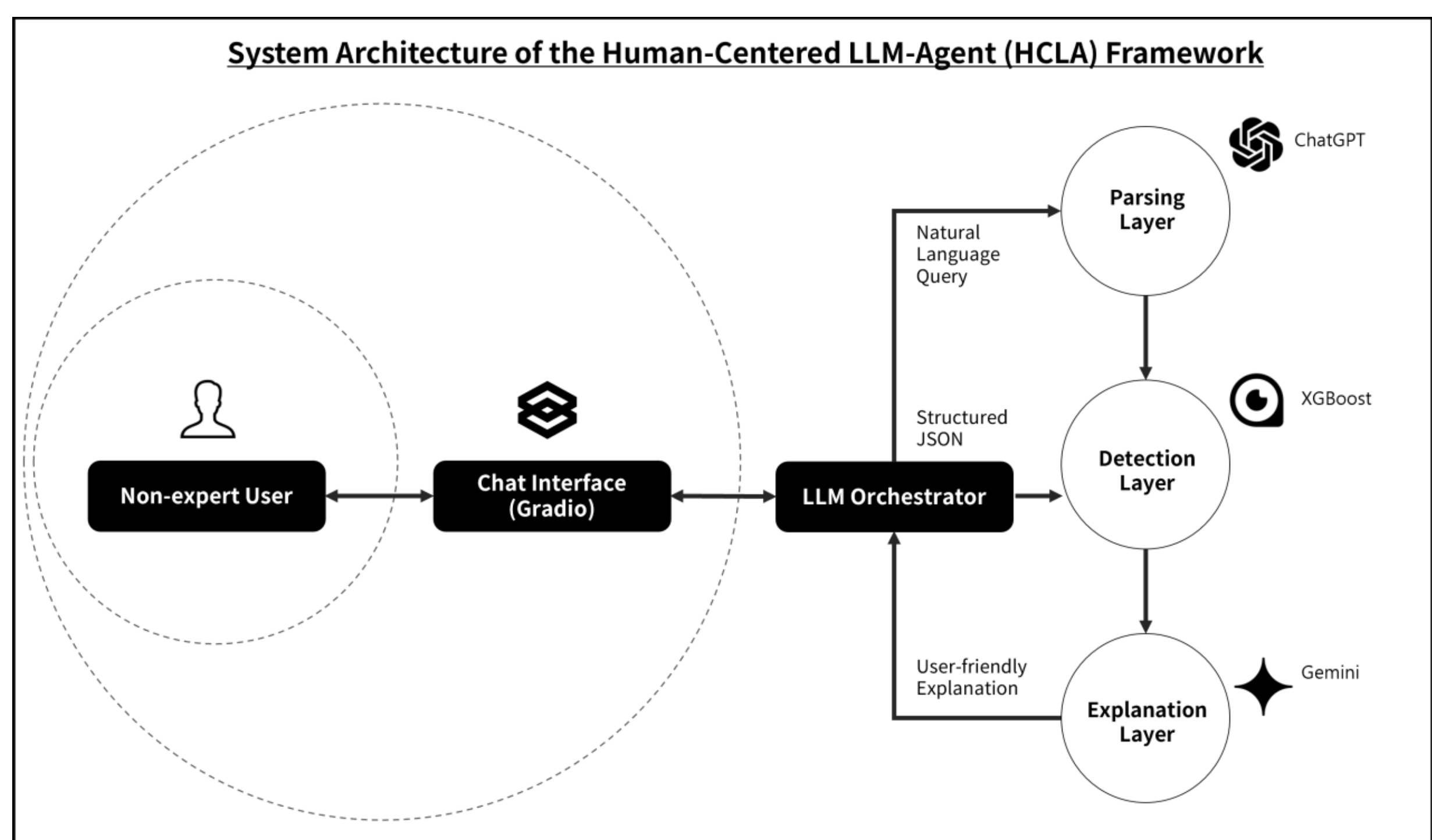

Figure 2: System Architecture of the Human-Centered LLM-Agent (HCLA) Framework.

Table 2: Performance Metrics of the Proposed System

| Metric | Value | Notes |
|---|---|---|
| Accuracy | 0.9159 | Overall correctness (XGBoost baseline) |
| Precision | 0.9317 | Avoids false positives |
| Recall | 0.9159 | Detects true anomalies |
| F1-score | 0.9209 | Harmonic mean of P/R |
| Latency (m/query) | $< 2$ | Avg. response time |
| Interpretability | Qualitative ↑ | LLM explanations improved comprehension |

prompts, the baseline detector achieves Accuracy 0.9159, Precision 0.9317, Recall 0.9159, F1 0.9209, with∼<2 m average response latency per query in the interactive loop.What the Human-Centered Layer Adds

Accessibility. Non-experts issue queries without writing filters or code; the system handles schema construction (see Listing 1–Listing 2).

Interpretability. Explanations tie scores to concrete patterns—frequency, counterparties, timing—rather than exposing raw numbers alone.

Trust Auditability. Each step (Parsing → Detection → Explanation) is visible and queryable, so users can check consistency and replay reasoning. (cf. Figure 1 –Figure 2 , Table tab:relatedwork.)

**Discussion Summary.** Results confirm that integrating LLM-based parsing and explanation layers maintains strong anomaly detection accuracy while enhancing interpretability and accessibility. Through modular orchestration, HCLA transforms anomaly detection from a static classification task into an interactive reasoning process—advancing the broader vision of *human-centered AI for financial transparency.*

## Simulated User Study

To approximate human-centered usability under limited resources, this study conducted a simulated user study with a Micro-Expert Panel (n=32) composed of experts in the fields of AI and Digital Assets. All participants held at least a master's degree in AI-related disciplines and possessed sufficient domain knowledge and professional experience to simulate real user judgment and reasoning processes.

The experiment was designed around three distinct anomalous transaction scenarios, each involving complex blockchain-based information such as transaction flows, input–output patterns, transaction values, and linked addresses. Participants compared and evaluated two types of explanatory formats for each case. The first format was the baseline model output XGBoost numerical dashboard, which presented only prediction results and probability values. The second format was a narrative explanation generated by HCLA, providing a natural-language interpretation and summary of the anomaly detection results (see Figure 4).

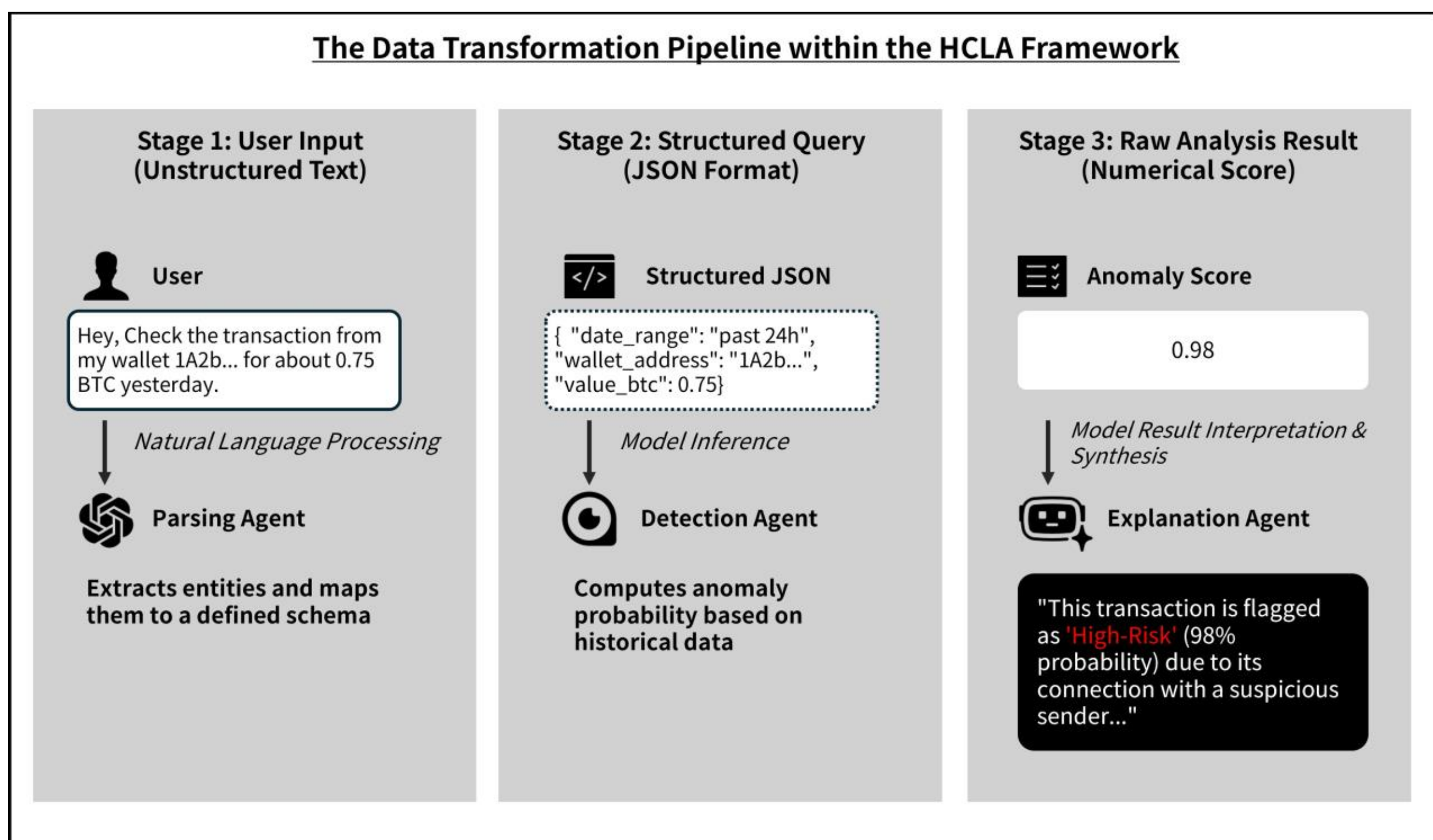

Figure 3: The Data Transformation Pipeline within the HCLA Framework.

After reviewing both explanations for each scenario, participants responded to survey items assessing *comprehension accuracy*, *trust*, and *clarity*. Comprehension was measured through two multiple-choice questions per case to evaluate both understanding of the presented scenario and the sincerity of participation (e.g., *"What action did the system ask the user to take?"* with choices: *(1) Verify the address and amount, and contact customer support, (2) Request transaction cancellation, (3) Change the password, (4) Ignore and proceed*). Trust and clarity were assessed using six and five items, respectively, rated on a 7-point Likert scale (1 = *strongly disagree* to 7 = *strongly agree*) (e.g., *"This system's results and explanations are consistent."* for trust, and *"This explanation is clear and satisfying."* for clarity). The elements were adapted from validated instruments in previous studies (Hoffman et al. 2018) measuring trust and explanatory clarity, and were modified to fit the experimental context of this investigation.

Internal consistency analysis showed that all constructs achieved high reliability, with Cronbach's $\alpha$ values exceeding 0.80 across all items. Specifically, the clarity dimension demonstrated extremely high reliability ($\alpha = 0.94$–$0.98$), while the trust dimension remained stable within the range of 0.82–0.90. These results indicate that all items consistently measured the intended constructs.

A paired t-test was conducted to compare perceived usability between the XGBoost and HCLA explanations. Across all three cases, participants rated the HCLA explanations significantly higher in both trust and clarity ($p < .001$). The negative $t$-values indicate that the mean ratings for HCLA were consistently higher than those for XGBoost, suggesting that the narrative explanations generated by HCLA had a more positive effect on users' understanding and confidence compared to purely numerical outputs.

In summary, this Micro-Expert Panel experiment empirically validated the effectiveness of the proposed AI explanation model in a simulated user environment. The results demonstrate that HCLA explanations outperform traditional numerical dashboards in terms of comprehensibility, trustworthiness, and clarity. Thus, this approach provides an efficient and practical methodological alternative for conducting human-centered evaluations of AI systems under limited resource conditions.

Table 3: Paired t-test results comparing XGBoost and HCLA explanations

| Anomaly Case | Measure | t | p-value |
|---|---|---|---|
| 1 | Trust | -5.586 | < .001 |
| 2 | Trust | -5.416 | < .001 |
| 3 | Trust | -4.629 | < .001 |
| 1 | Clarity | -5.865 | < .001 |
| 2 | Clarity | -6.542 | < .001 |
| 3 | Clarity | -5.417 | < .001 |

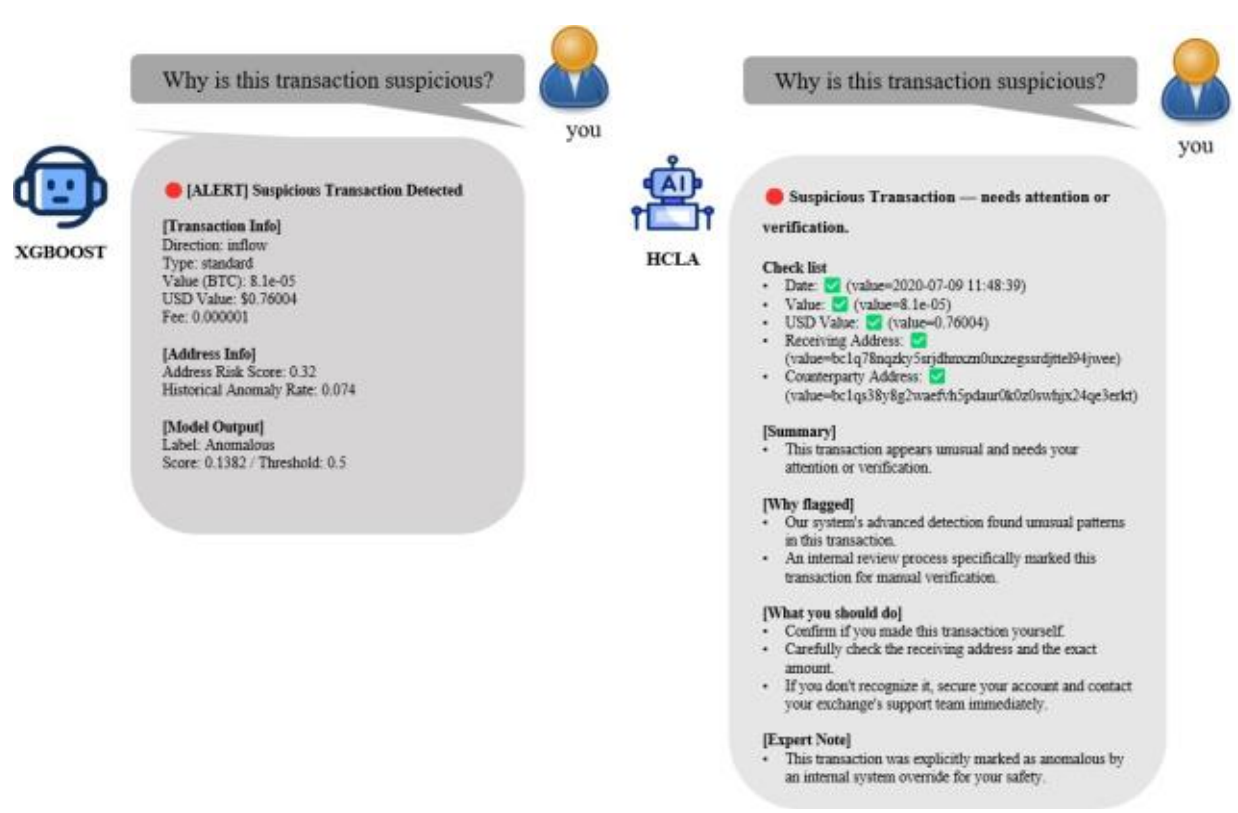

Figure 4: XGBoost vs. HCLA explanations for a suspicious transaction (Anomaly Case 1).

## Experimental Results

The experimental evaluation of the **HCLA** framework focused on assessing its effectiveness from a *user-centered perspective*, rather than solely on model-level performance. While traditional anomaly detection studies emphasize accuracy or recall, this research evaluates how the integration of LLM-based agents enhances accessibility, interpretability, and user trust in digital asset forensics.

### Human-Centered Design Advantages

The HCLA system demonstrates several benefits that align with the principles of **human-centered AI**—bringing transparency, adaptability, and inclusivity into complex analytical workflows.

**Accessibility.** Conventional anomaly detection tools require technical proficiency, specialized data preparation, and prior knowledge of blockchain semantics. In contrast, HCLA enables users to interact through natural-language dialogue using the Gradio interface. Even non-technical participants were able to submit queries such as:

> *"Analyze transactions from my wallet over the past week and highlight anything suspicious."*

The system autonomously parsed, structured, and analyzed this input without requiring manual schema creation. This conversational workflow significantly lowers the barrier to entry for analysts, auditors, and general users, allowing them to explore data without coding or model configuration.

**Interpretability.** Explainability emerged as one of the strongest advantages of the proposed framework. Unlike black-box detection models that output numerical anomaly scores, HCLA provides **context-rich narrative explanations**. The *Explainer Agent (Gemini)* translated each model output into intuitive feedback, e.g.:

> *"The transaction shows moderate anomaly due to unusually high frequency of small-value transfers to unknown counterparties."*

Such linguistic grounding enables users to understand not only *what* was flagged as anomalous but also *why*. Through this interpretive layer, users reported higher confidence in the system's reasoning and decision support.

**Trust and Transparency.** A recurring issue in financial AI systems is the opacity of decision boundaries. The multi-agent structure of HCLA explicitly addresses this by decomposing reasoning steps: the **Parsing Agent** converts language into structured data, the **Detection Agent** executes a conventional ML-based classification, and the **Explainer Agent** reconstructs the reasoning process in human terms. By making each stage visible and queryable, users can "see through" the pipeline rather than blindly trusting a single output. This transparency builds trust, which is essential for deployment in regulatory, compliance, and audit contexts.

## Discussion

### Identified Limitations

Despite its success in improving usability and interpretability, several limitations were identified through experimental analysis and user testing:

- **Computational Cost and Latency:** The use of LLMs introduces computational overhead, resulting in a certain level of response latency. This may pose limitations in real-time monitoring scenarios or high-frequency transaction streams.
- **Domain-Specific Adaptation:** General-purpose LLMs such as ChatGPT and Gemini occasionally produced ambiguous interpretations or inconsistent terminology (e.g., mixing "cluster" and "wallet" references). Domain-specific fine-tuning with blockchain and finance corpora is expected to improve reasoning accuracy and reduce linguistic drift.
- **Scalability Constraints:** Although the current prototype performs effectively on batch-processed datasets, scalability to continuous blockchain streams will require asynchronous orchestration and caching mechanisms.

These findings highlight that while the HCLA framework achieves interpretability and accessibility, optimization of computational efficiency and contextual accuracy remains an ongoing challenge.

### Future Research Directions

**1) Human-Centered Advantages.** HCLA advances human-centered AI by: Accessibility: Natural-language interaction removes coding or schema knowledge barriers. Interpretability: Narrative reasoning grounds numeric outputs in domain semantics. Trust Transparency: Users can trace logic across Parsing–Detection–Explanation stages, fostering verifiable accountability.

**2) Limitations.** While effective, limitations remain: Computational Latency: LLM calls introduce 2–3 s delay per query, limiting real-time monitoring. Domain Adaptation: Generic LLMs occasionally confuse blockchain terminology; fine-tuning is planned. Sample Generality: The micro-expert panel's small academic sample limits external validity.

**3) Future Directions.** Domain-Specific Fine-Tuning: Adapt LLMs with finance-corpus knowledge to reduce semantic drift. Real-Time Extension: Integrate streaming pipelines (Kafka/Flink) for continuous detection. Multimodal Expansion: Fuse text, screenshots, and logs for cross-modal reasoning. Large-Scale User Validation: Conduct IRB-approved studies to generalize trust and interpretability findings.

## Conclusion

The accelerating complexity of digital asset ecosystems demands anomaly detection systems that are not only technically advanced but also human-centered and interpretable. This paper introduced the **Human-Centered LLM-Agent (HCLA)** framework—a multi-agent architecture that integrates large language models with graph-informed XGBoost analytics to detect anomalous cryptocurrency transactions in an accessible and transparent manner. By embedding human–AI collaboration into the detection process, the framework redefines how non-expert users can interact with and interpret high-dimensional financial data.

Unlike conventional rule-based or black-box machine learning models, the proposed system decomposes detection into three cognitively aligned stages:

(i) **Parsing Agent**, which translates unstructured natural-language queries into structured feature representations;

(ii) **Detection Agent**, which executes anomaly classification through an XGBoost model augmented with LLM reasoning; and

(iii) **Explanation Agent**, which converts probabilistic outputs into context-rich, conversational narratives that support human decision-making.

This modular and interpretable pipeline creates a continuous feedback loop between users and models, improving transparency, adaptability, and trustworthiness. The human-centered architecture demonstrated reduced cognitive burden for non-technical stakeholders, who could understand anomaly causes and implications through guided conversational analysis.

**Core contributions** of this research can be summarized as follows:

- Introduction of an integrated LLM-Agent and XGBoost architecture for digital asset anomaly detection, enabling interpretable and scalable detection workflows.
- Design of an automated conversational pipeline that unifies natural-language parsing, structured model inference, and contextual explanation.
- Empirical demonstration that a multi-agent LLM design enhances user accessibility, interpretability, and engagement compared to static or single-agent baselines.

**Future research** will advance the HCLA paradigm in several directions. First, we aim to incorporate *real-time monitoring* and continuous learning pipelines that dynamically adapt to evolving market behaviors. Second, integrating *external multimodal data sources*—including logs, screenshots, or news streams—can strengthen contextual anomaly reasoning. Third, domain-specific fine-tuning of LLM components will be explored to improve accuracy, reduce latency, and enhance alignment with financial semantics. Finally, we envision deploying the HCLA framework as an open, auditable platform for AI-assisted financial forensics, contributing to a broader agenda of trustworthy and human-aligned AI in digital finance.

Ultimately, this work bridges the gap between algorithmic intelligence and human sensemaking in digital asset forensics. By situating large language models not as mere assistants but as orchestrators of analytic workflows, the proposed HCLA framework illustrates a scalable pathway toward explainable, adaptive, and user-first anomaly detection in complex financial ecosystems.

## Acknowledgments

This work was supported by the National Research Foundation of Korea (NRF) grant funded by the Korean government under the project “Socio-Technological Solutions for Bridging the AI Divide: A Blockchain and Federated Learning-Based AI Training Data Platform” (NRF-2024S1A5C3A02043653).

## References

Devireddy, A.; and Huang, S. 2025. CALM: A Framework for Continuous, Adaptive, and LLM-Mediated Anomaly Detection in Time-Series Streams. Dataflow ML Team Technical Report. Available upon request or internal distribution.

He, X.; Wu, D.; Zhai, Y.; and Sun, K. 2025. SentinelAgent: Graph-based Anomaly Detection in LLM-based Multi-Agent Systems. https://arxiv.org/abs/2502.12345. Manuscript under review, arXiv:2502.12345.

Hoffman, R. R.; Mueller, S. T.; Klein, G.; and Litman, J. 2018. Metrics for explainable AI: Challenges and prospects. *arXiv preprint arXiv:1812.04608*.

Jia, Y.; Wang, Y.; Sun, J.; Tian, Y.; and Qian, P. 2025. LMAE4Eth: Generalizable and Robust Ethereum Fraud Detection by Exploring Transaction Semantics and Masked Graph Embedding. ArXiv preprint, arXiv:2509.03939.

Lei, Y.; Xiang, Y.; Wang, Q.; Dowsley, R.; Yuen, T. H.; Choo, K.-K. R.; and Yu, J. 2025. Large Language Models for Cryptocurrency Transaction Analysis: A Bitcoin Case Study. ArXiv preprint arXiv:2501.18158, arXiv:2501.18158.

Li, Y.; Hu, J.; Hooi, B.; He, B.; and Chen, C. 2025. DGP: A Dual-Granularity Prompting Framework for Fraud Detection with Graph-Enhanced LLMs. arXiv:2505.12346.

Lin, Y.; Tang, J.; Zi, C.; Zhao, H. V.; Yao, Y.; and Li, J. 2025. UniGAD: Unifying Multi-level Graph Anomaly Detection. ArXiv preprint, arXiv:2503.12345.

Park, T. 2024. Enhancing Anomaly Detection in Finan- cial Markets with an LLM-Based Multi-Agent Framework. ArXiv preprint arXiv:2403.19735v1, arXiv:2403.19735.

Russell-Gilbert, A. 2025. *RAAD-LLM: Adaptive Anomaly Detection Using LLMs and RAG Integration*. Ph.d. dissertation, Mississippi State University. ProQuest Document ID: 31934376.

Sun, J.; Jia, Y.; Wang, Y.; Tian, Y.; and Zhang, S. 2025. Ethereum Fraud Detection via Joint Transaction Language Model and Graph Representation Learning. *Information Fusion*, 111: 103074. Available online, February 2025.

Tsai, C.-P.; Teng, G.; Wallis, P.; and Ding, W. 2025. ANOLLM: Large Language Models for Tabular Anomaly Detection. In *Proceedings of the International Conference on Learning Representations (ICLR)*. Singapore: OpenReview. Published as a conference paper at ICLR 2025.

Watson, A. 2024. *Explain First, Trust Later: LLM-Augmented Explanations for Graph-Based Crypto Anomaly Detection*. Master's thesis, School of Engineering Technology, Purdue University, West Lafayette, United States. Contact: watso213@purdue.edu.

Watson, A. 2025. Explain First, Trust Later: LLM-Augmented Explanations for Graph-Based Crypto Anomaly Detection. arXiv:2504.14567.

Yu, J.; Wu, X.; Liu, H.; Guo, W.; and Xing, X. 2025. BlockScan: Detecting Anomalies in Blockchain Transactions. arXiv:2505.12345.